\documentclass{article}

\usepackage[preprint, nonatbib]{neurips_2026}

\usepackage[utf8]{inputenc}
\usepackage[T1]{fontenc}
\usepackage{hyperref}
\usepackage{url}
\usepackage{booktabs}
\usepackage{amsfonts}
\usepackage{amsmath}
\usepackage{amssymb}
\usepackage{nicefrac}
\usepackage{microtype}
\usepackage{xcolor}
\usepackage{graphicx}
\usepackage{svg}
\usepackage{multirow}
\usepackage{listings}
\usepackage{wrapfig}
\usepackage{caption}
\usepackage[backend=biber,style=numeric,sorting=none]{biblatex}
\providecommand{\Description}[1]{}

\title{Structured Layout Priors for Robust Out-of-Distribution Visual Document Understanding}

\author{%
  Peter El Hachem \\
  ETH Zurich \\
  Rämistrasse, Switzerland\\
  \texttt{pelhachem@ethz.ch}
  \And
  Ahmed Nassar \\
  IBM Research \\
  Rüschlikon, Switzerland \\
  \texttt{ahn@zurich.ibm.com}
  \And
  A. Said Gurbuz \\
  IBM Research \\
  Rüschlikon, Switzerland \\
  \texttt{said.gurbuz@ibm.com}
  \AND
  Christoph Auer \\
  IBM Research \\
  Rüschlikon, Switzerland \\
  \texttt{cau@zurich.ibm.com}
  \And
  Peter W. J. Staar \\
  IBM Research \\
  Rüschlikon, Switzerland \\
  \texttt{taa@zurich.ibm.com}
}
\begin{document}

\maketitle

\begin{abstract}
Vision-Language Models (VLMs) parse documents end-to-end but frequently break down on layouts unlike those seen in training. We attribute this to a two-hop bottleneck: before the decoder can extract content (Hop 2), it must first classify and localize the enclosing layout entity (Hop 1), and when the first hop fails the second collapses into omissions, malformed structure, or autoregressive repetition. We pre-resolve Hop 1 outside the decoder by running a lightweight RT-DETR detector, serializing its outputs in the parser's native DocTags vocabulary, and injecting them into the prompt alongside the full page image. Unlike analyze-then-parse approaches that crop the page, or prior prompt-level priors written in plain text, our prior shares the decoder's generation space and leaves the global image in view as a fallback when detections are noisy. On a 10k-page structural out-of-distribution benchmark, markdown F1 rises from $0.37$ to $0.92$; on the Chinese subset of OmniDocBench, table TEDS rises from $0.01$ to $0.36$; and on the 26k-page ViDoRe V3 benchmark, infinite-loop decoding failures drop across every industrial domain tested. These gains cost $15\%$ wall-clock latency and a median of $74$ prompt tokens, with no architectural change to the base VLM. An attention-level analysis further reveals a bimodal phase shift in which the decoder attends to injected layout tokens when emitting structure and to image patches when emitting content, consistent with the two-hop bottleneck being alleviated. Model weights will be released to support reproducibility.
\end{abstract}

\section{Introduction}
Document understanding remains a central challenge in computer vision because most information is stored in visually rich, unstructured formats  \cite{baviskar2021invoice}. Recent Vision-Language Models (VLMs) significantly simplify OCR pipelines by jointly modeling vision and text, enabling end-to-end conversion from page images to structured markup \cite{kim2022Donut, blecher2023nougatneuralopticalunderstanding, ibm2025docling}. However, structured document generation creates a {\bfseries structural dependency}: the decoder must first identify a region's boundaries and class, then emit its content within the enclosing markup, typically as an interleaved pattern such as $\langle\mathrm{tag}\rangle\;\mathrm{content}\;\langle/\mathrm{tag}\rangle$. Recent systems already hint at this dependency by using region grounding or coordinate-based prompting to stabilize parsing \cite{wei2025deepseekocrcontextsopticalcompression,wang2024qwen2vlenhancingvisionlanguagemodels,wei2024generalocrtheoryocr20}, yet the mechanism behind these gains has remained largely implicit.

We make this dependency explicit by framing end-to-end document parsing as a two-hop problem: first, \textit{entity resolution}, where visual features are localized and mapped to a layout class; second, \textit{content extraction}, where the decoder parses the identified region. When the first hop fails, errors propagate to the second and manifest as omissions, malformed structure, or repetitive decoding, and this effect is amplified in compact models whose decoder capacity is already constrained. The two-hop view has recently been studied for factual recall in multimodal models \cite{cohen2026performancegapentityknowledge,constantin2025two-hop-problem}; we adapt it here to structured document generation.

Building on this framing, we pre-resolve the first hop outside the autoregressive loop. A lightweight RT-DETR detector \cite{zhao2024detrsbeatyolosrealtime} predicts document regions and their coordinates, and we serialize these detections as location-tag priors before appending them to the task prompt. The decoder thus receives a symbolic page map alongside the full page image, and can allocate its capacity to Hop 2 rather than recovering layout structure from pixels. Unlike analyze-then-parse pipelines that feed detector-defined crops to the recognizer \cite{feng2025dolphindocumentimageparsing,duan2026glmocrtechnicalreport}, we keep the full page in view so the decoder retains a global fallback when the detector is imperfect.

Our key contributions are as follows:

(i) \textbf{A layout-conditioned fine-tuning pipeline}: We introduce a ``Guided-Decoding'' framework that injects structured RT-DETR layout priors directly into the VLM prompt. This decouples layout localization from content extraction, effectively targeting the ``two-hop'' bottleneck without requiring architectural changes.

(ii) \textbf{Mechanistic evidence for capacity reallocation}: Through attention analysis, we reveal a \textit{bimodal attention phase shift}: the decoder attends to layout tokens for structural syntax and to image patches for textual content. This is consistent with the pipeline actively resolving the first hop, freeing the compact decoder's capacity for content extraction.

(iii) \textbf{Robust out-of-distribution (OOD) generalization}: Across severe structural, linguistic, and complexity shifts, our method raises markdown F1 from $0.37$ to $0.92$ on NoveltySet, table TEDS from $0.01$ to $0.36$ on the Chinese subset of OmniDocBench, and reduces infinite-loop decoding failures on every industrial domain of ViDoRe V3. Ablations confirm the prior acts as lightweight, flexible guidance, adding only $15\%$ latency rather than a rigid constraint.

\section{Related Work}

\paragraph{Document VLMs} Traditional OCR pipelines decompose document extraction into text detection, recognition, and downstream parsing \cite{ocr++2016ocrrobustframeworkinformation, ocr-llm-india2025digitizationdocumentinformationextraction}. These systems often accumulate error across stages and lose spatial semantics when layouts are complex or noisy \cite{rijhwani-etal-2020-ocr, ramirezorta2022postocrdocumentcorrectionlarge}. Recently, the field has shifted toward \textit{generative} transformer-based architectures, evolving into modern solutions based on \textit{Vision-Language Models (VLMs)}. These models adopt joint training paradigms for layout and content parsing while removing the dependency on external OCR engines. Frameworks such as Donut \cite{kim2022Donut}, Nougat \cite{blecher2023nougatneuralopticalunderstanding}, DeepSeek-OCR \cite{wei2025deepseekocrcontextsopticalcompression}, Qwen2-VL \cite{wang2024qwen2vlenhancingvisionlanguagemodels}, and recent document-expert variants like SmolDocling \cite{nassar2025smoldoclingultracompactvisionlanguagemodel} unify document extraction into a single-stage decoding task.

\paragraph{Layout conditioning and structural priors} The diversity of documents poses a challenge in developing a unified model that can comprehend intricate relationships between text and visual objects across a wide range of document types, formats, and tasks. Researchers have pursued several complementary strategies to bridge the gap between visual and text signals. Early approaches like \textit{LayoutLM} \cite{LayoutLM_2020, xu2022layoutlmv2multimodalpretrainingvisuallyrich, huang2022layoutlmv3} sought to bridge this gap by incorporating 2D positional embeddings. These embeddings represent coordinates $(x_0, y_0, x_1, y_1)$ of each bounding box rather than individual text tokens, allowing the model to treat the document as a spatial map rather than a linear string. However, these models require an external OCR engine, meaning system performance is strictly capped by the quality of the OCR.

In the generative era, methods to align vision and language have evolved. Cross-attention blocks, introduced by \textit{Flamingo} \cite{alayrac2022flamingovisuallanguagemodel}, and self-attention linear projections, used by LLaVA-based models \cite{LLaVA2024}, successfully map visual features to language models but often struggle with downstream tasks that necessitate fine-grained visual groundedness or high-resolution spatial details. To provide better structural signposting, \textit{SmolVLM} \cite{marafioti2025smolvlmredefiningsmallefficient} introduces the use of \emph{media introduction tokens}. However, these markers only signal the presence of visual content without conveying its internal structure.

More recent works frame layout conditioning as an analyze-then-parse pipeline. \textit{Dolphin} \cite{feng2025dolphindocumentimageparsing} and \textit{GLM-OCR} \cite{duan2026glmocrtechnicalreport} adopt a two-stage pipeline with precomputed layout analysis. However, a key difference is that both parse detector-defined regions as separate crops, making the downstream parser tightly dependent on detector quality: missed boxes remove content entirely, and inaccurate boxes corrupt the local crop seen by the recognizer. In our approach, the detector provides only a structural prior, while the full page image is still passed to the vision encoder. As a result, the model retains global visual context and can still use the original page evidence even when the injected layout prior is imperfect. 

Most closely related to our work, Zhu et al.~\cite{zhu2025simpleeffectivelayouttoken} inject layout tokens as a plain-text prior to a general-purpose VLM. Our contribution differs in three respects: (i) our priors are serialized in the parser's own DocTags vocabulary, not plain text, aligning with the decoder's generation space; (ii) we target a 258M-parameter document-expert VLM where the two-hop bottleneck is most acute; and (iii) we provide attention-level evidence for the mechanism by which layout priors help, leading to improved stability and performance.

\begin{table}[ht]
\caption{Comparison of detector-conditioned document parsers.
``Crops'' = recognizer sees detector-defined sub-images only.
``Full page'' = vision encoder sees the entire page, with detections
used only as a prompt-level prior. ``Native'' = prior tokens come from
the parser's own tokenizer.}
\label{tab:positioning}
\centering
\small
\begin{tabular}{lcccc}
\toprule
Method & Detector$\to$Prompt & Page image input & Prior tokens & Scale \\
\midrule
Dolphin~\cite{feng2025dolphindocumentimageparsing} & yes & crops & plain text & 8B \\
GLM-OCR~\cite{duan2026glmocrtechnicalreport} & yes & crops & mixed & $\sim$1.3B \\
Zhu et al.~\cite{zhu2025simpleeffectivelayouttoken} & yes & full page & plain text & general-purpose \\
Qwen2-VL~\cite{wang2024qwen2vlenhancingvisionlanguagemodels} & endogenous & full page & native & 2--72B \\
\textbf{Ours} & yes & \textbf{full page} & \textbf{native (DocTags)} & \textbf{0.258B} \\
\bottomrule
\end{tabular}
\end{table}

\paragraph{Autoregressive collapse in document parsers}
A significant limitation in transformer-based document parsers is \textbf{autoregressive degradation}, frequently manifesting as generation collapse. In its simplest form, the last sentence or paragraph is repeated over and over again. This phenomenon is a well-documented failure mode in neural text generation, particularly when employing greedy decoding strategies \cite{holtzman2020curiouscaseneuraltext, blecher2023nougatneuralopticalunderstanding, duan2026glmocrtechnicalreport}. Typically, collapse is triggered by an initial localized error, such as a misinterpretation of a dense visual artifact, from which the model’s internal hidden states are unable to recover. As the sequence length increases, the attention mechanism becomes increasingly biased toward the recently generated repetitive tokens, leading to a total loss of semantic progression.

Our method injects layout priors directly in the native \textit{DocTags} syntax, integrating naturally with the decoder's token-level predictions (e.g., \textit{<text><loc\_100><loc\_200><loc\_300><loc\_400></text>}). By doing so, we counteract generative instability by anchoring the model's attention. We formalize this two-step decomposition later in the pipeline description in Section~\ref{sec:two-stage-pipeline}.

Taken together, the literature shows that layout conditioning helps, but existing solutions either crop the page and inherit any detector error, inject the prior as plain text that is foreign to the decoder's vocabulary, or operate at a scale where the two-hop bottleneck is easy to absorb. Our work targets the remaining gap: a compact, DocTags-native, full-page-preserving prior whose mechanism we make observable through attention-level analysis.

\section{Methodology}
\subsection{Implementation details}
\paragraph{Base Model} \textit{granite-docling-258M} is used as the base VLM for our fine-tuning pipeline. It combines a \textit{google/siglip2-base-patch16-512} vision backbone \cite{tschannen2025siglip2multilingualvisionlanguage}, which encodes document images up to 2048$\times$2048 pixels via dynamic patching, with an Idefics3-style pixel-shuffle connector and a compact \textit{Granite-165M} decoder that autoregressively generates structured DocTags outputs.

\paragraph{Layout detector} We utilize \textit{docling-layout-heron}, an RT-DETR-based detector that achieves high-fidelity localization of document elements (mAP@50:95 = 0.69 on DocLayNet \cite{Pfitzmann_2022, livathinos2025advancedlayoutanalysismodels}). By leveraging an end-to-end transformer framework, it delivers significant precision gains over traditional YOLO-based systems \cite{redmon2016lookonceunifiedrealtime} while maintaining the low-latency inference necessary for real-time document processing. The model outputs bounding boxes in a normalized top-left coordinate system. Detected object categories are detailed in Appendix C, Table~\ref{tab:heron_class_map}.

\paragraph{Markup Language}\textit{DocTags} is a structured markup language that represents document objects and their bounding-box coordinates as tokens \cite{nassar2025smoldoclingultracompactvisionlanguagemodel}. To pre-resolve Hop 1, we map precomputed layout objects into DocTags format before injecting them into the prompt. A complete list of DocTags is provided in Appendix C Table~\ref{tab:doctag_vocabulary}.

\begin{figure*}
\begin{center}
\includesvg[width=\linewidth]{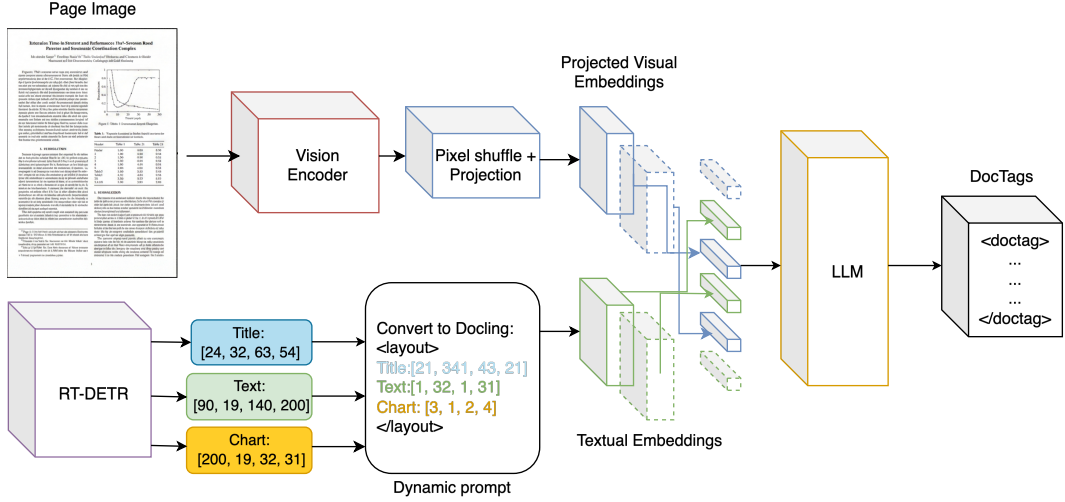}
\end{center}
\caption{Overview of our two-stage pipeline}
\Description{Overview of our two-stage pipeline. Stage 1 runs a lightweight layout detector to precompute document objects and coordinates. Stage 2 serializes these detections into DocTags and injects them into the VLM prompt to support structure-aware decoding.}
\label{fig:two-stage-pipeline}
\end{figure*}

\subsection{Two-Stage Parsing Pipeline}
\label{sec:two-stage-pipeline}
\paragraph{Stage 1: Layout Precomputation} Stage 1 augments the existing VLM training and inference loops without any change to the annotation format: as each image is loaded, \textit{docling-layout-heron} \cite{livathinos2025advancedlayoutanalysismodels} produces a set of layout detections on the fly. The detector was pretrained on a mixture that overlaps with the document distributions we care about, including DocLayNet \cite{Pfitzmann_2022} and  WordScape \cite{weber2023wordscapepipelineextractmultilingual}.

Let
\[
\mathcal{B} = \{(c_k, x_k^{\min}, y_k^{\min}, x_k^{\max}, y_k^{\max})\}_{k=1}^{K},
\]
where $c_k$ is the semantic class and the coordinates are normalized to $[0, 500]$. We serialize this set into a DocTags layout prior $L = \mathrm{serialize}(\mathcal{B})$, which conditions generation as
\[
p_\psi(R \mid I) \approx p_\psi(R \mid I, L), \qquad L \sim p_\phi(\cdot \mid I),
\]
where $\phi$ denotes the detector parameters and $R$ is the target parse. This factorization separates structural proposal from autoregressive content generation: Stage 1 commits to candidate regions and their classes, and the decoder conditions on both the image and the serialized blueprint when generating the parse. Resolving the class-plus-location decision at the prompt level pre-answers Hop 1, so the decoder's autoregressive capacity can be spent on Hop 2 (content extraction).

\paragraph{Stage 2: VLM Parsing} Given the pre-solved structural layout, the second stage performs multimodal synthesis. The input image is processed via a Vision Encoder, which adaptively partitions the image into a dynamic number of patches based on the aspect ratio. Each patch is mapped to a 64-token visual representation. Passing the full page image preserves a visual fallback when detections are incomplete, at the cost of duplicated information (detections describe
regions that the encoder already sees). Section \ref{sec:ablations} shows this redundancy is useful: the model remains functional under high detection dropout, suggesting it treats the prior as guidance rather than groundtruth.

To bridge the modality gap, these visual tokens are projected through a Multi-Layer Perceptron (MLP) into the shared embedding space of the language model. The sequence of visual tokens and the DocTag-augmented prompt are then fed into the LLM decoder. By conditioning the linguistic generation on both the raw visual features and the validated structural "blueprint," the model generates a layout-enriched parse of the document with higher fidelity than ungrounded baselines.

\subsection{Training Strategy}
\paragraph{\bfseries Phase 1: Pretraining}
The pretraining phase jointly optimizes two objectives: \textit{modality alignment} and \textit{document understanding}. For modality alignment, we train the connector (a pixel-shuffle projector) to map visual features from the SigLIP-2 backbone into the embedding space of the Granite-165M language model using the \textit{Cauldron} dataset \cite{laurencon2024mattersbuildingvisionlanguagemodels}. This stage emphasizes dense-text pages with relatively simple layouts so the model first learns reliable character recognition before tackling harder structural reasoning.

For document understanding, we then expose the model to a hybrid corpus that combines synthetic examples produced by the Docling ensemble pipeline (e.g., TableFormer outputs) with high-quality human-labeled annotations. The manually verified subset covers complex structures such as nested tables, mathematical formulas, and multi-column flows, providing supervision for richer document parsing.

The model is optimized using a standard autoregressive next-token prediction objective \cite{liu2023visualinstructiontuning}. Given an input image $I$ and a text instruction $P$, the model minimizes the negative log-likelihood of the target sequence $R$. By segmenting the output tokens into label, location, and content types, the model is trained to jointly perform object classification, spatial localization, and content extraction within a unified sequence modeling framework.

\paragraph{\bfseries Phase 2: Layout Injection}
Building upon the pretrained structured checkpoint, Phase 2 introduces a structural paradigm shift by transitioning the model from a "one-shot" parser to a {\bfseries guided two-stage architecture}. For each training batch, we employ docling-layout-heron to detect page elements. We apply post-processing to retain high-confidence detections (score $>0.6$), merge fragmented regions, and perform non-maximum suppression to handle overlapping boxes. These detections are then serialized into the \textit{DocTags} format. By injecting these serialized detections as an explicit structural prior, we decompose the parsing task into two distinct steps: (i) \textit{entity resolution} and (ii) \textit{content extraction}. Stage 1 (Heron) pre-resolves the ''where" and ''what" by providing the layout class and spatial coordinates in the prompt. Conditioned on this blueprint, the decoder can dedicate its full capacity to ''Hop 2", extracting and generating content from the visual evidence rather than exhausting parameters on spatial discovery.

\paragraph{\bfseries Phase 3: Guided Fine-tuning}
In this final phase, we concentrate gradient on the tokens whose prediction actually requires multimodal reasoning. Because location tokens are copied verbatim from the prompt to the output, supervising them teaches the model only to copy, so we mask them out of the loss. We keep the loss on label tokens: they dictate the structural context, and any error in their generation propagates into the enclosed content and undermines the pipeline's reliability. Concretely, we use a \textbf{Masked Cross-Entropy Loss} with an indicator function $M(x_i)$ that suppresses location tokens ($\texttt{<loc\_0>}, \dots, \texttt{<loc\_500>}$) during backpropagation:

\begin{equation}
    \mathcal{L}_{guided} = -\sum_{i=1}^{|R|} M(x_i) \cdot \log p_\psi(x_i \mid I, P, x_{<i})
\end{equation}

where:
\begin{equation}
    M(x_i) = 
    \begin{cases} 
      0 & \text{if } x_i \in \mathcal{T}_{location} \\
      1 & \text{otherwise}
    \end{cases}
\end{equation}
and $\mathcal{T}_{location}$ is the set of all spatial coordinate tokens.

Training, optimization, and detector hyperparameters are summarized in Appendix~\ref{app:hyperparameters}, Table~\ref{tab:hyperparameters}.

\section{Evaluation}
\label{sec:evaluation}

\begin{figure*}[t]
\centering
\begin{minipage}[t]{0.45\textwidth}
\centering
\includesvg[height=0.26\textheight,keepaspectratio]{assets/stability_vidore_horizontal.svg}
\captionof{figure}{Stability benchmarking on ViDoRe dataset for infinite loops across 7 different industries. Red bars are the base model, green bars are the finetuned model.}
\Description{Horizontal bar plot comparing base model to the finetuned model for the percentage of infinite loops across industries within ViDoRe dataset}
\label{fig:ViDoRe-stability}
\end{minipage}\hfill
\begin{minipage}[t]{0.52\textwidth}
\centering
\includesvg[height=0.26\textheight,keepaspectratio]{assets/tsne_2d_with_vidore.svg}
\captionof{figure}{t-SNE visualization of training data, NoveltySet, and selected ViDoRe subsets. Quantitative MMD details and SVM separability results are reported for the comparison between training data and NoveltySet in Appendix Table~\ref{tab:mmd_summary}, Appendix Figure~\ref{fig:svm_roc_appendix}, and Table~\ref{tab:svm_scores_appendix}.}
\Description{Vision embeddings computed using shared ViT encoder for all datasets (train, NoveltySet and ViDoRe ). Figure shows a t-SNE qualitative separability in 2 dimensions, using a perplexity set to 30.}
\label{fig:ood_sep}
\end{minipage}
\end{figure*}

Our core hypothesis is that layout-conditioned fine-tuning transfers better to out-of-distribution (OOD) documents than standard end-to-end decoding. We evaluate along three complementary OOD axes and, for completeness, an in-distribution benchmark:

\begin{itemize}
  \item \textbf{Structural shift:} NoveltySet, a proprietary 10k-page human-annotated benchmark curated to exhibit page structures and object compositions absent from training.
  \item \textbf{Linguistic shift:} the Chinese subset of the public OmniDocBench~\cite{ouyang2025omnidocbenchbenchmarkingdiversepdf}.
  \item \textbf{Stability under complexity:} 26k pages of the public ViDoRe V3~\cite{loison2026vidorev3comprehensiveevaluation} spanning seven industrial domains.
  \item \textbf{In-distribution (reference):} DocLayNet~\cite{Pfitzmann_2022}.
\end{itemize}

For DocLayNet and NoveltySet, reference annotations are available in DocTags format, enabling direct structural and content evaluation with \textit{docling-eval}. For OmniDocBench, we follow its official evaluation protocol, which scores model outputs from generated Markdown files.

\paragraph{OOD Evaluation: Language Generalization} Out of the full training mixture used for both basemodel \textit{granite-docling-258M} and finetuned \textit{granite-docling-2stage-258M}, around 1\% of training data is Chinese documents. To evaluate this generalization, we benchmark both models on OmniDocBench \cite{ouyang2025omnidocbenchbenchmarkingdiversepdf}, which contains 1355 PDF images as mixture of English and Chinese documents. To assess the generalization across language fairly, we benchmark our models by filtering out English documents, keeping only Chinese documents.

\begin{figure}[t]
\centering
\includesvg[width=0.7\linewidth]{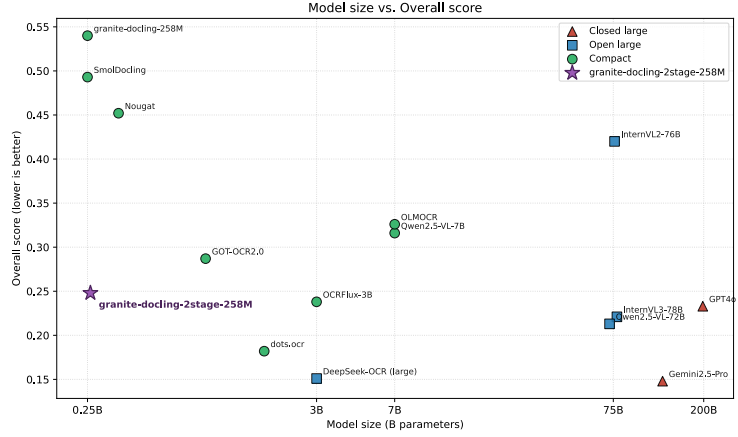}
\caption{Pareto comparison on the English subset of OmniDocBench. The plot compares edit-distance performance across state-of-the-art document parsers as a function of model size, highlighting that \textit{granite-docling-2stage-258M} remains competitive at 0.25B parameters.}
\label{fig:omnidocbench_pareto}
\Description{Pareto scatter plot comparing document parsing models on the English subset of OmniDocBench, with model size on one axis and edit-distance performance on the other. The figure highlights the position of granite-docling-2stage-258M among larger state-of-the-art baselines.}
\end{figure}

The results show while text-level metrics can remain similar or slightly degrade, structure-sensitive objects significantly improve. Especially for table extraction, we can see an improvement from near 0\% to \textit{43\%}. 
This pattern suggests that layout-prior injection is particularly effective for structure-centric parsing such as table extraction, regardless of the language of the underlying content. To contextualize this result, we also benchmark our models on the English subset of OmniDocBench using edit-distance metrics across text, table, formula, and reading-order extraction, and compare them against a range of state-of-the-art specialist and general-purpose document parsers. The Pareto comparison in Figure~\ref{fig:omnidocbench_pareto} shows that \textit{granite-docling-2stage-258M} remains competitive despite its compact 0.25B parameter scale. Full multilingual results are reported in Table~\ref{tab:omnidocbench_multilingual}, and the complete English-only comparison is reported in Appendix Table~\ref{tab:omnidocbench_multilingual_all_models}.

\begin{table*}
\caption{OmniDocBench benchmark results for \textit{granite-docling-258M} vs \textit{granite-docling-2stage-258M} across Chinese (ZH) and English (EN) subsets. \textbf{ED}: Edit Distance,
\textbf{TEDS}: Tree Edit Distance-based Similarity,
\textbf{TEDS-S}: TEDS Structure only,
\textbf{RO}: Reading Order.}
\label{tab:omnidocbench_multilingual}
    \setlength{\tabcolsep}{6pt} 
    \resizebox{\textwidth}{!}{%
    \begin{tabular}{llccccccc}
        \toprule
        & & \multicolumn{2}{c}{\textbf{Text}} & \multicolumn{3}{c}{\textbf{Table}} & \textbf{Formula} & \textbf{RO} \\
        \cmidrule(lr){3-4} \cmidrule(lr){5-7} \cmidrule(lr){8-8} \cmidrule(lr){9-9}
        \textbf{Lang.} & \textbf{Model} & \textbf{ED} $\downarrow$ & \textbf{BLEU} $\uparrow$ & \textbf{TEDS} $\uparrow$ & \textbf{TEDS-S} $\uparrow$ & \textbf{ED} $\downarrow$ & \textbf{ED} $\downarrow$ & \textbf{ED} $\downarrow$ \\
        \midrule
        \multirow{2}{*}{ZH} & granite-docling-258M & 0.84 & \textbf{0.04} & 0.01 & 0.01 & 0.98 & 0.86 & 0.73 \\
        & \textbf{granite-docling-2stage-258M} & \textbf{0.62} & 0.02 & \textbf{0.36} & \textbf{0.43} & \textbf{0.57} & \textbf{0.84} & \textbf{0.52} \\
        \midrule
        \multirow{2}{*}{EN} & granite-docling-258M & 0.78 & 0.10 & 0.43 & 0.47 & 0.44 & \textbf{0.29} & 0.69 \\
        & \textbf{granite-docling-2stage-258M} & \textbf{0.17} & \textbf{0.48} & \textbf{0.47} & \textbf{0.52} & \textbf{0.34} & 0.30 & \textbf{0.19} \\
        \bottomrule
    \end{tabular}%
    }
\end{table*}

\paragraph{OOD Evaluation: Stability under complexity}\label{sec:exp_stability}
To quantify the impact of structural prompting on decoding robustness, we evaluate our model on the \textit{ViDoRe V3} dataset \cite{loison2026vidorev3comprehensiveevaluation}, encompassing 26,000 pages across diverse industrial domains. We use \textit{ViDoRe V3} for its industrial diversity and structural complexity, and assess robustness independently of ground-truth availability. Using a high-throughput vLLM inference engine, we monitor generation for timeout-inducing infinite loops and define a failure as any sequence exceeding a maximum token threshold ($T_{max}=5000$) without EOS termination. Results are qualitatively unchanged for $T_{\max} \in \{2000, 10000\}$. We report the ratio of infinite loops across industries for the base model \textit{granite-docling-258M} and the finetuned model \textit{granite-docling-2stage-258M}.

\paragraph{Findings.} Conditioning generation on layout priors yields more stable decoding across all domains (Figure~\ref{fig:ViDoRe-stability}). Without structural grounding, the VLM more frequently fails on dense visual regions, complex infographics, and schematic diagrams, entering repetitive autoregressive loops. The t-SNE analysis (Figure~\ref{fig:ood_sep}) supports this pattern: subsets that occupy distinct regions of the feature space relative to the training distribution, including HR, Energy, and Finance (EN), exhibit the highest baseline failure rates, and Finance (FR) does as well, which is consistent with sensitivity to a language shift given the limited French-language training data. The linguistic generalization aligns with our earlier findings: gains are driven primarily by structured elements such as tables, where DocTags provide an unambiguous spatial scaffold regardless of the underlying language. Quantitatively, the mean infinite-loop rate drops from $3.9\%$ to $2.4\%$ across the seven industrial domains, with the largest absolute reductions on the most OOD subsets (HR: $7.4\%{\to}3.6\%$; Finance EN: $4.6\%{\to}1.5\%$; Energy: $4.3\%{\to}2.2\%$); CS, already at $0.9\%$, is unchanged within noise.

\paragraph{OOD evaluation: Structural generalization} NoveltySet is a proprietary dataset of over 10000 human annotated images, created within the Docling team.
In our evaluation protocol, we treat NoveltySet as \textbf{out-of-distribution (OOD)} with respect to the \textbf{entire training dataset}, as it exhibits different page structures and object compositions.

To quantify this shift, we compute image-level representations for both training-set images and NoveltySet images using the frozen, fine-tuned vision encoder of \textit{granite-docling-2stage-258M}. For an image $x_i$, let $P_i$ be the dynamic number of visual patches produced by the encoder. Each patch is represented by 64 tokens of dimension 576, so we denote patch-token features as $V_i \in \mathbb{R}^{P_i \times 64 \times 576}$ with elements $V_{i,p,t}$. We define the pooled embedding by averaging over all patch tokens:
\[
z_i = \frac{1}{64P_i}\sum_{p=1}^{P_i}\sum_{t=1}^{64} V_{i,p,t} \in \mathbb{R}^{576}.
\]
Using these embeddings, we quantify ID/OOD separation in two ways. First, we train a binary linear SVM to classify dataset labels (training data vs.~NoveltySet), obtaining a near-perfect F1 score of 0.99. Second, we compute Maximum Mean Discrepancy (MMD) with an RBF kernel, yielding an unbiased MMD value of $0.512$.

Full MMD kernel details are reported in Appendix Table~\ref{tab:mmd_summary}, and SVM ROC and classification metrics are reported in Appendix Figure~\ref{fig:svm_roc_appendix} and Table~\ref{tab:svm_scores_appendix}. Together, these results, along with the t-SNE visualization in Figure~\ref{fig:ood_sep}, show that NoveltySet occupies a distinct region in feature space, supporting its use as an OOD benchmark relative to the full training distribution. Having established NoveltySet as OOD, we benchmark produced Markdown text using the same metrics as DocLayNet. Results are reported in Table~\ref{tab:merged_model_performance}.

\paragraph{In Distribution evaluation} DocLayNet \cite{Pfitzmann_2022} is a human-annotated dataset with 80,863 unique pages, resized to $1024\times1024$, spanning six document categories. In this section, we compare Markdown extraction performance across \textit{DeepSeek-OCR} \cite{wei2025deepseekocrcontextsopticalcompression}, \textit{smol-docling} \cite{nassar2025smoldoclingultracompactvisionlanguagemodel}, \textit{granite-docling-258M}, and our fine-tuned model \textit{granite-docling-2stage-258M}. Results are reported in Table~\ref{tab:merged_model_performance}. \textit{granite-docling-2stage-258M} matches or exceeds every in-distribution baseline on every metric, and in particular improves over the un-conditioned \textit{granite-docling-258M} backbone across the board (BLEU $0.65{\to}0.70$, F1 $0.84{\to}0.85$, edit distance $0.29{\to}0.27$), indicating that the layout prior does not trade in-distribution accuracy for out-of-distribution robustness.

\begin{table*}
\caption{Model performance on DocLayNet (ID) and NoveltySet (OOD). The two-stage model closes most of the OOD gap left by the baseline, raising F1 from $0.37$ to $0.92$ and edit distance from $0.78$ to $0.39$ on NoveltySet, while remaining competitive in-distribution.}
\label{tab:merged_model_performance}
\centering
\setlength{\tabcolsep}{6pt}
\resizebox{\textwidth}{!}{%
\begin{tabular}{llcccccc}
    \toprule
    \textbf{Setting} & \textbf{Model} & \textbf{BLEU} $\uparrow$ & \textbf{F1-Score} $\uparrow$ & \textbf{Precision} $\uparrow$ & \textbf{Recall} $\uparrow$ & \textbf{Meteor} $\uparrow$ & \textbf{Edit-Dist.} $\downarrow$ \\
    \midrule
    \multirow{4}{*}{ID} 
    & smol-docling & 0.59 & 0.80 & 0.89 & 0.78 & 0.67 & 0.48 \\
    & DeepSeek-OCR & 0.61 & 0.82 & 0.88 & 0.80 & 0.77 & 0.29 \\
    & granite-docling-258M & 0.65 & 0.84 & 0.91 & 0.83 & 0.72 & 0.29 \\
    & \textbf{granite-docling-2stage-258M} & \textbf{0.70} & \textbf{0.85} & \textbf{0.92} & 0.83 & \textbf{0.79} & \textbf{0.27} \\
    \midrule
    \multirow{2}{*}{OOD} 
    & granite-docling-258M & 0.10 & 0.37 & 0.94 & 0.26 & 0.25 & 0.78 \\
    & \textbf{granite-docling-2stage-258M} & \textbf{0.62} & \textbf{0.92} & \textbf{0.95} & \textbf{0.90} & \textbf{0.83} & \textbf{0.39} \\
    \bottomrule
\end{tabular}%
}
\end{table*}

\section{Experiments}

\label{sec:experiments}
We evaluate the proposed layout-injection framework with three complementary experiments: (1)
Mechanistic interpretability of attention behavior, (2) Ablations for training,
and (3) overhead of precomputation.

\paragraph{Experiment 1: Mechanistic Interpretability} \label{sec:exp_mech} To evaluate the efficacy of structural prompt injection in mitigating recall errors, we conduct a mechanistic interpretability study on the model's internal attention mechanisms. Specifically, we investigate the allocation of the decoder's computational budget, quantified via attention weights, across visual embeddings and injected structural tokens during autoregressive generation. This analysis has been conducted on a single NVIDIA L4 GPU.
For each document page, we perform a forward pass and extract the attention matrices from the language decoder at each decoding step $t$.

Following the standard scaled dot-product attention \cite{vaswani2023attentionneed}, the attention weight matrix $A$ for a given layer and head is defined as:\begin{equation}A = \mathrm{softmax}\left(\frac{QK^\top}{\sqrt{d_k}}\right)\end{equation}To synthesize these signals across all layers $\mathcal{L}$ and heads $\mathcal{H}$, we employ a max-pooling aggregation:\begin{equation}\mathrm{Agg}(i,j) = \max_{l \in \mathcal{L}, h \in \mathcal{H}} A_{l,h}[i,j]\end{equation}The choice of the $\max$ operator over the arithmetic mean is strategic. While the mean may dilute sparse, high-magnitude attention signals, the $\max$ operator effectively isolates the primary "anchors" (e.g., coordinate tags or specific image patches) that exert the strongest influence on the current token's hidden state.

Our analysis reveals a distinct \textbf{bimodal attention shift}. When the decoder generates structural syntax (e.g., layout or coordinate tokens), the peak attention mass aligns predominantly with the injected layout tokens. Conversely, during the generation of textual content, attention reverts to the corresponding image patches and preceding context. This phase-shift is consistent with the two-hop decomposition: the injected layout tokens function as spatial priors that ground the VLM's Hop 1 entity resolution in the precomputed structural metadata, which would otherwise have to be recovered from pixels during decoding. A complementary per-token attention-distribution plots for generated layout tags and location tokens are provided in Appendix Figures~\ref{fig:layout_tag_attention_distribution_appendix} and~\ref{fig:loc_tag_attention_distribution_appendix}.

\paragraph{Experiment 2: Ablation Studies and Robustness Analysis} \label{sec:ablations} A two-stage pipeline introduces a potential error propagation pathway, where inaccuracies in the layout precomputation stage may degrade downstream generation. To quantify this sensitivity, we conduct a multi-axis ablation on the \textbf{DocLayNet} test split~\cite{Pfitzmann_2022}, evaluating three perturbations: \textbf{Permutation}, comparing reading-order injection against randomized ordering, \textbf{Stochastic Injection}, varying injection probability ($p \in \{0.8, 1.0\}$) to probe whether parsing capability is preserved when priors are absent and \textbf{Partial Masking}, applying a dropout rate of 0.3 over layout components to simulate missed detections. The results show that the VLM is a robust consumer of noisy layout metadata, maintaining high fidelity under all three perturbations. Full ablation results are provided in Appendix~\ref{app:appendix_ablation}, Table~\ref{tab:ablation_markdown_extraction}.

\paragraph{Experiment 3: Layout Precomputation Overhead} We evaluate the computational cost of the proposed two-stage pipeline by benchmarking the inference runtime on the \textit{DocLayNet} \cite{Pfitzmann_2022} test split using an NVIDIA A100 GPU. Excluding I/O operations, the \texttt{docling-layout-heron} detector executes at a median latency of only 28~ms per page. When integrated into the full pipeline, \texttt{granite-docling-2stage-258M} achieves a median processing speed of 2.6~seconds per page, representing a marginal $15\%$ overhead compared to the $2.26$~seconds per page required by the baseline \texttt{granite-docling-258M}. In our experiments, this latency increase coincides with improved structural recall and decoding stability (see Section~\ref{sec:exp_stability}). Full cross-hardware runtime statistics, including performance on consumer-grade hardware, are provided in Appendix~\ref{app:appendix_c_runtime_heron}

Moreover, we also measure the number of layout tokens injected into the prompt on the DocLayNet test split. Benchmarked on DocLayNet, a median of 74 tokens are added from the precomputation. Corresponding appendix details are provided in Appendix~\ref{app:appendix_c_tokens}, Table~\ref{tab:heron-token-overhead}.

\section{Conclusion and Limitations}
We demonstrated that decoupling layout localization from content extraction significantly improves the out-of-distribution generalization of compact VLMs. By offloading layout analysis to a specialized detector and injecting explicit spatial anchors as prompt priors, our approach stabilizes decoding on dense, complex documents. This two-stage paradigm adds only $15\%$ latency and a median of $74$ prompt tokens, suggesting a practical path toward more reliable document parsing with compact VLMs.

While promising, our findings have limitations. First, the pipeline relies on the upstream layout detector: although our ablations show the VLM is tolerant to moderate noise, arbitrarily poor structural proposals can yield misleading priors, and gains are not guaranteed to transfer uniformly to document families, scripts, or visual styles absent from our evaluation. Second, the two-stage paradigm introduces a practical systems trade-off between modest latency and the added detector, serialization, and prompt-budget cost, which may constrain ultra-strict deployments. Finally, because our study focuses on empirical performance and stability, the mechanisms by which structural priors improve generalization should be viewed as evidence-backed hypotheses rather than complete causal explanations.



\printbibliography
\clearpage
\appendix
\section{Training Hyperparameters}
\label{app:hyperparameters}

Table~\ref{tab:hyperparameters} summarizes the model, optimization, and detector settings used throughout all training phases.

\begin{table*}[t]
\centering
\small
\caption{Training hyperparameters for the proposed two-stage document parser.}
\label{tab:hyperparameters}
\setlength{\tabcolsep}{8pt}
\begin{tabular}{l l}
\toprule
\textbf{Hyperparameter} & \textbf{Value} \\
\midrule
Vision encoder & SigLIP-2 \\
Language model & Granite-165M \\
Multimodal connector & Pixel-shuffle projector / MLP \\
Optimizer & AdamW \\
Learning-rate schedule & Linear warmup (500 steps) + cosine decay \\
Backbone learning rate $\eta_{\text{backbone}}$ & $2 \times 10^{-5}$ \\
Projector learning rate $\eta_{\text{mp}}$ & $5.12 \times 10^{-4}$ \\
Gradient clipping & Max norm $1.0$ \\
Per-device batch size & 2 \\
Gradient accumulation steps & 8 \\
Effective batch size & 1,024 \\
Hardware & 64$\times$ NVIDIA H100 80GB \\
Total training steps & 244,140 across phases $T_1$, $T_2$, and $T_3$ \\
Phase-3 training loss & Masked cross-entropy excluding location tokens \\
Detector confidence threshold $\tau$ & 0.6 \\
Detector NMS IoU threshold $\iota$ & 0.5 \\
Detector post-processing & Confidence filtering, fragment merging, and NMS \\
\bottomrule
\end{tabular}
\end{table*}

\section{Ablation Study Results}
\label{app:appendix_ablation}

Table~\ref{tab:ablation_markdown_extraction} reports the ablation results for structural augmentation on the \textbf{DocLayNet} \cite{Pfitzmann_2022} test split.

\begin{table*}
\caption{Performance comparison across various structural augmentation parameters. Configurations follow the naming convention [S]-[P]-[D]: S indicates shuffling (ys) or in-order injection (ns), P denotes the probability of layout injection per sample, and D represents the dropout ratio, where a random subset of detected components is discarded for a percentage of samples.}
\label{tab:ablation_markdown_extraction}
\centering
\small
\begin{tabular}{lcccccc}
\hline
Configuration & BLEU & F1-Score & Precision & Recall & METEOR & Edit-Dist \\
\hline
ns - 0.8 - 0.0 & 0.684 & 0.850 & 0.907 & 0.819 & 0.766 & 0.276 \\
ns - 1.0 - 0.0 & 0.680 & 0.851 & 0.909 & 0.816 & 0.770 & 0.274 \\
ys - 0.8 - 0.0 & 0.677 & 0.846 & 0.903 & 0.812 & 0.761 & 0.280 \\
ys - 1.0 - 0.0 & 0.690 & 0.854 & 0.913 & 0.822 & 0.772 & 0.273 \\
ys - 1.0 - 0.3 & 0.691 & 0.856 & 0.913 & 0.825 & 0.774 & 0.271 \\
\hline
\end{tabular}
\end{table*}

\section{Out-of-Distribution Shift Validation}
This appendix summarizes the quantitative checks used to support our treatment of NoveltySet as an out-of-distribution benchmark relative to the training distribution. We report two complementary diagnostics: a kernel-based two-sample test over pooled vision embeddings and a linear separability analysis based on dataset-label prediction. Together, these results provide additional context for the OOD evaluation presented in Section~\ref{sec:evaluation}.

\subsection{MMD heuristic and kernel settings for ID/OOD separation}
\label{app:mmd_details}

Table~\ref{tab:mmd_summary} reports the kernel configuration and resulting MMD values used in Section~\ref{sec:evaluation}.

\begin{table}
\caption{MMD summary for training-data (ID) vs NoveltySet (OOD) using pooled vision embeddings.}
\label{tab:mmd_summary}
\centering
\small
\setlength{\tabcolsep}{6pt}
\begin{tabular}{l r}
\toprule
Metric & Value \\
\midrule
Feature dimension & 576 \\
$\sigma$ (median heuristic) & 82.984 \\
$\gamma=1/(2\sigma^2)$ & $7.26074\times10^{-5}$ \\
$\mathrm{MMD}^2_{\mathrm{biased}}$ & 0.262944 \\
$\mathrm{MMD}^2_{\mathrm{unbiased}}$ & 0.262272 \\
$\mathrm{MMD}_{\mathrm{biased}}$ & 0.512781 \\
$\mathrm{MMD}_{\mathrm{unbiased}}$ & 0.512125 \\
\bottomrule
\end{tabular}
\end{table}

\subsection{SVM ROC and classification scores for ID/OOD separability}
\label{app:svm_scores}

\begin{figure}
\centering
\includesvg[width=0.65\linewidth]{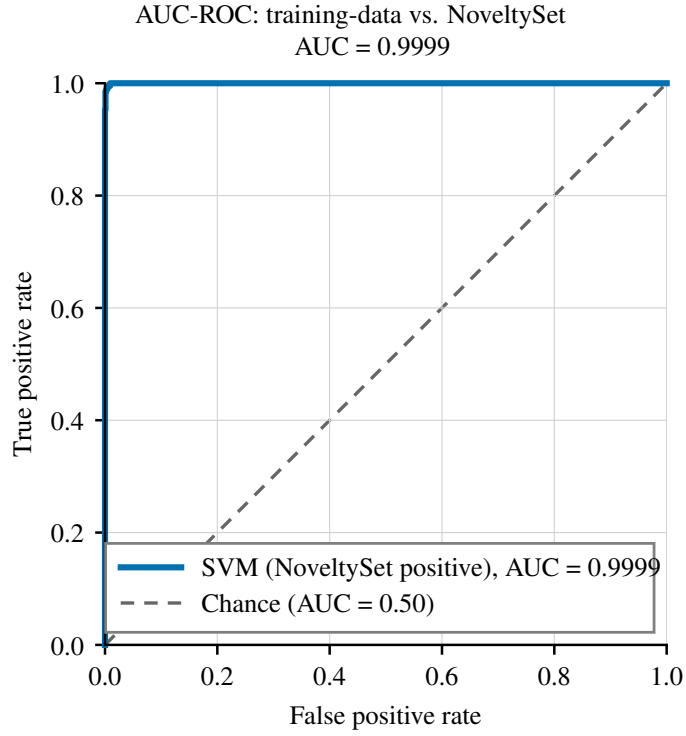}
\caption{Linear SVM ROC curve for training-data vs NoveltySet separability.}
\label{fig:svm_roc_appendix}
\Description{AUC for varying thresholds, showing high separability of data}
\end{figure}

Figure~\ref{fig:svm_roc_appendix} reports the ROC curve of the linear SVM used for dataset-label separation, and Table~\ref{tab:svm_scores_appendix} reports the corresponding classification scores on DocLayNet.

\begin{table}
\caption{Linear SVM evaluation scores on DocLayNet.}
\label{tab:svm_scores_appendix}
\centering
\small
\setlength{\tabcolsep}{8pt}
\begin{tabular}{l c}
\toprule
Metric & Value \\
\midrule
Precision & 1.0000 \\
Recall & 0.9748 \\
F1 & 0.9872 \\
\bottomrule
\end{tabular}
\end{table}

\subsection{Mechanistic attention heatmap}
\label{app:mech_heatmap}

\begin{figure}[t]
\centering
\includesvg[width=0.9\linewidth]{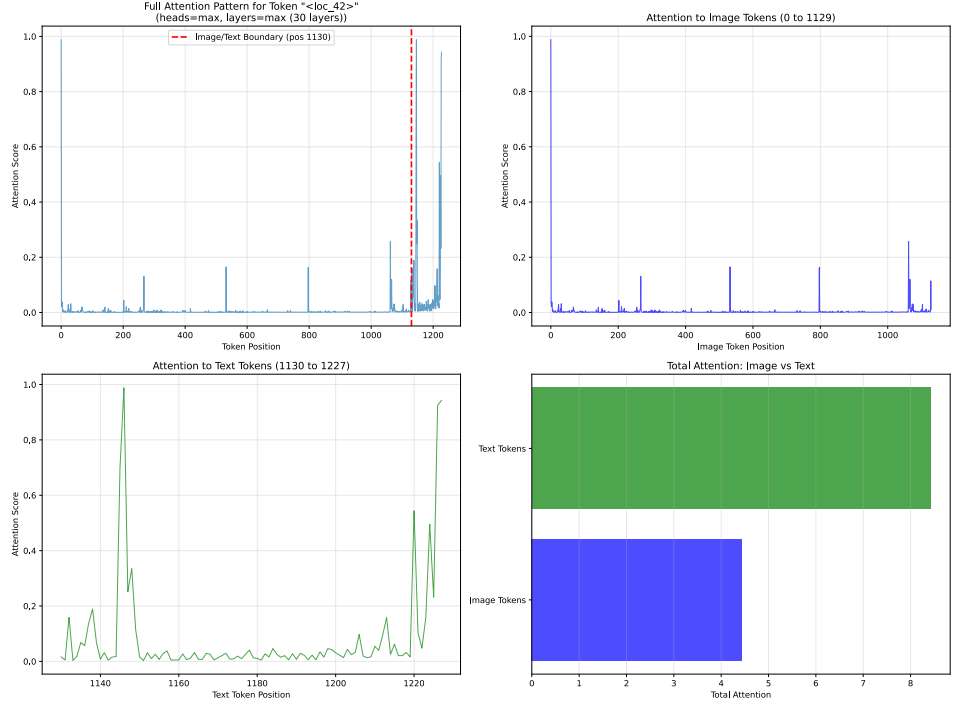}
\caption{Attention distribution conditioned on generating a location token. The mass concentrates on injected structural tokens, consistent with the decoder relying on prompt-level layout priors while emitting document structure.}
\label{fig:layout_tag_attention_distribution_appendix}
\Description{Attention distribution plot for layout-tag generation, showing concentration over injected structural tokens.}
\end{figure}

\begin{figure}[t]
\centering
\includesvg[width=0.9\linewidth]{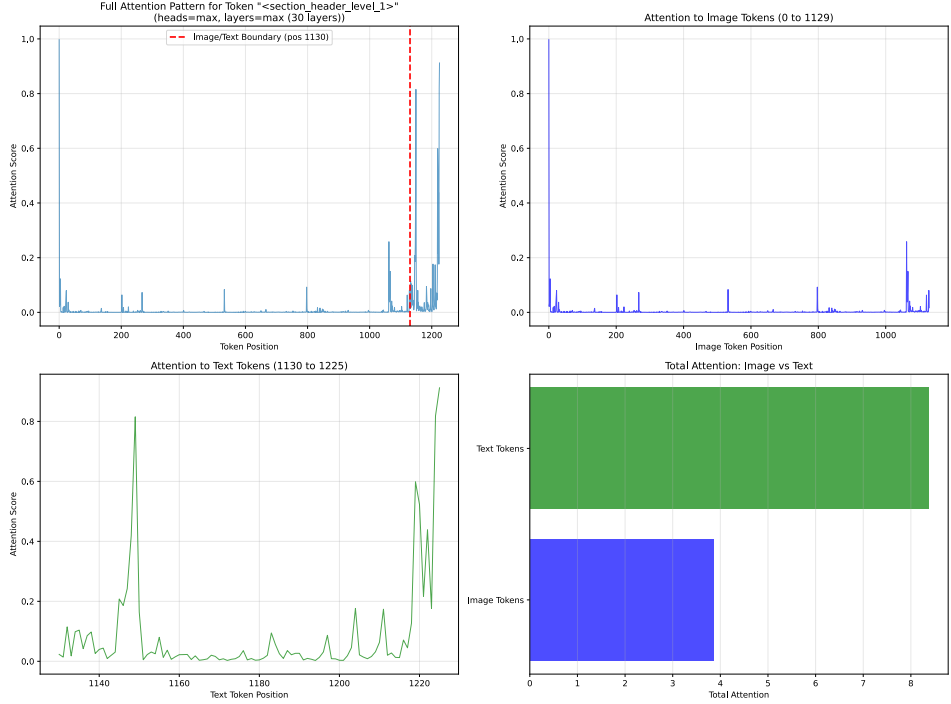}
\caption{Attention distribution conditioned on generating a layout token. The distribution remains concentrated on structural prompt tokens, consistent with the decoder using serialized coordinates to resolve spatial structure before returning to image-grounded content extraction.}
\label{fig:loc_tag_attention_distribution_appendix}
\Description{Attention distribution plot for location-token generation, showing concentration over structural prompt tokens.}
\end{figure}

Figures~\ref{fig:layout_tag_attention_distribution_appendix} and~\ref{fig:loc_tag_attention_distribution_appendix} further break this behavior down by generated token type. They report the attention distribution when the decoder emits a layout tag and when it emits a location token, respectively, making the bimodal shift discussed in Section~\ref{sec:exp_mech} explicit.

\subsection{OmniDocBench English-only comparison}
\label{app:omnidocbench_english}

Table~\ref{tab:omnidocbench_multilingual_all_models} reports the full English-only OmniDocBench comparison referenced in Section~\ref{sec:evaluation}, including larger specialist and general-purpose baselines.

\begin{table*}
\caption{English-only OmniDocBench comparison across state-of-the-art models. Metrics are edit distance ($\downarrow$) for text, table, formula, and reading order.}
\label{tab:omnidocbench_multilingual_all_models}
\setlength{\tabcolsep}{6pt}
\resizebox{\textwidth}{!}{%
\begin{tabular}{llccccc}
\toprule
\textbf{Lang.} & \textbf{Model} & \textbf{Size} & \textbf{Text} & \textbf{Table} & \textbf{Formula} & \textbf{RO} \\
\midrule
\multirow{2}{*}{ZH}
& granite-docling-258M & 0.25B & 0.84 & 0.98 & 0.86 & 0.73 \\
& \textbf{granite-docling-2stage-258M} & 0.25B & \textbf{0.62} & \textbf{0.57} & \textbf{0.84} & \textbf{0.52} \\
\midrule
\multirow{16}{*}{EN}
& GPT4o & $\pm$200B & 0.144 & 0.234 & 0.425 & 0.128 \\
& Gemini2.5-Pro & 128B & 0.055 & \textbf{0.130} & 0.356 & \textbf{0.049} \\
& InternVL3-78B & 78B & 0.117 & 0.279 & 0.380 & 0.095 \\
& InternVL2-76B & 76B & 0.353 & 0.547 & 0.543 & 0.317 \\
& Qwen2.5-VL-72B & 72B & 0.092 & 0.341 & 0.315 & 0.106 \\
& Dolphin & 8B & 0.352 & 0.465 & 0.258 & 0.35 \\
& Qwen2.5-VL-7B & 7B & 0.151 & 0.598 & 0.376 & 0.138 \\
& olmOCR & 7B & 0.097 & 0.608 & 0.455 & 0.145 \\
& OCRFlux-3B & 3B & 0.112 & 0.269 & 0.447 & 0.126 \\
& DeepSeek-OCR (Large) & 3B & 0.054 & 0.277 & 0.152 & 0.067 \\
& dots.ocr & 1.7B & 0.137 & 0.166 & 0.320 & 0.182 \\
& GOT-OCR2.0 & 0.9B & 0.189 & 0.459 & 0.360 & 0.141 \\
& Nougat & 0.35B & 0.365 & 0.572 & 0.488 & 0.382 \\
& granite-docling-258M & 0.25B & 0.78 & 0.44 & 0.31 & 0.69 \\
& SmolDocling & 0.25B & 0.262 & 0.729 & 0.753 & 0.227 \\
\midrule
& \textbf{granite-docling-2stage-258M} & 0.25B & 0.17 & 0.34 & \textbf{0.29} & 0.19 \\
\bottomrule
\end{tabular}%
}
\end{table*}

\newpage
\section{The docling-layout-heron detector}

\subsection{Class list}
\label{app:appendix_c_class_list}

The \textit{docling-layout-heron} detector predicts 17 layout classes, listed in Table~\ref{tab:heron_class_map}.

\begin{table}
\centering
\small
\caption{Heron layout-object class map used in this work.}
\label{tab:heron_class_map}
\setlength{\tabcolsep}{8pt}
\begin{tabular}{c l}
\toprule
Class ID & Class Name \\
\midrule
0 & Caption \\
1 & Footnote \\
2 & Formula \\
3 & List-item \\
4 & Page-footer \\
5 & Page-header \\
6 & Picture \\
7 & Section-header \\
8 & Table \\
9 & Text \\
10 & Title \\
11 & Document Index \\
12 & Code \\
13 & Checkbox-Selected \\
14 & Checkbox-Unselected \\
15 & Form \\
16 & Key-Value Region \\
\bottomrule
\end{tabular}
\end{table}

\subsection{DocTags}
DocTags constitute a subset of the model's expanded tokenizer, representing document structure. As detailed in Table~\ref{tab:doctag_vocabulary}, these tags are categorized into: (i) \textbf{Layout Tags}, which provide semantic classification for document regions, and (ii) \textbf{Location Tags}, which define spatial coordinates via a discrete grid of 501 quantized tokens.

\begin{table}[htbp]
    \centering
    \small
    \caption{DocTag Vocabulary: Specialized tokens integrated into the \textit{granite-docling-258M} tokenizer. We introduce \texttt{<otsl>} to represent Optimized Table Structure Language for hierarchical cell extraction.}
    \label{tab:doctag_vocabulary}
    \setlength{\tabcolsep}{12pt}
    \begin{tabular}{l l}
        \toprule
        \textbf{Tag Category} & \textbf{Tokens / Range} \\
        \midrule
        \textbf{Layout Tags} & \texttt{<page\_header>}, \texttt{<section\_header>}, \\
                             & \texttt{<unordered\_list>}, \texttt{<ordered\_list>}, \\
                             & \texttt{<list\_item>}, \texttt{<text>}, \texttt{<otsl>}, \\
                             & \texttt{<picture>}, \texttt{<page\_break>}, \\
                             & \texttt{<formula>}, \texttt{<code>}, \texttt{<page\_footer>} \\
        \midrule
        \textbf{Location Tags} & \texttt{<loc\_0>} \dots \texttt{<loc\_500>} \\
        \bottomrule
    \end{tabular}
\end{table}

\subsection{Mapping Heron output to DocTags}
\label{app:heron_mapping}

This subsection illustrates how raw layout detections are converted into the DocTags sequence used by \textit{granite-docling-258M}.

{\bfseries Original detections from Heron.}
\begin{lstlisting}
section-header: [49, 42, 197, 51]
text: [132, 56, 432, 71]
list-item: [132, 73, 421, 81]
list-item: [132, 82, 188, 90]
text: [131, 98, 246, 105]
picture: [132, 106, 440, 257]
text: [132, 269, 443, 291]
caption: [132, 296, 295, 302]
otsl: [131, 302, 441, 459]
page-footer: [48, 471, 183, 478]
\end{lstlisting}

{\bfseries Serialized DocTags prompt fragment.}
\begin{lstlisting}[breaklines=true, basicstyle=\small\ttfamily, columns=flexible]
<section_header><loc_49><loc_42><loc_197><loc_51></section_header>
<text><loc_132><loc_56><loc_432><loc_71></text>
<list_item><loc_132><loc_73><loc_421><loc_81></list_item>
<list_item><loc_132><loc_82><loc_188><loc_90></list_item>
<text><loc_131><loc_98><loc_246><loc_105></text>
<picture><loc_132><loc_106><loc_440><loc_257></picture>
<text><loc_132><loc_269><loc_443><loc_291></text>
<caption><loc_132><loc_296><loc_295><loc_302></caption>
<otsl><loc_131><loc_302><loc_441><loc_459></otsl>
<page_footer><loc_48><loc_471><loc_183><loc_478></page_footer>
\end{lstlisting}

{\bfseries Final prompt injected.}
\begin{lstlisting}
Convert this page to Docling:
<layout>
<section_header><loc_49><loc_42><loc_197><loc_51></section_header>
<text><loc_132><loc_56><loc_432><loc_71></text>
<list_item><loc_132><loc_73><loc_421><loc_81></list_item>
<list_item><loc_132><loc_82><loc_188><loc_90></list_item>
<text><loc_131><loc_98><loc_246><loc_105></text>
<picture><loc_132><loc_106><loc_440><loc_257></picture>
<text><loc_132><loc_269><loc_443><loc_291></text>
<caption><loc_132><loc_296><loc_295><loc_302></caption>
<otsl><loc_131><loc_302><loc_441><loc_459></otsl>
<page_footer><loc_48><loc_471><loc_183><loc_478></page_footer>
</layout>
\end{lstlisting}

\subsection{Runtime Overhead}
\label{app:appendix_c_runtime_heron}
Runtime performance of \textit{docling-layout-heron} in 3 hardware configurations: a single AMD EPYC
7763 64-Core, a single Nvidia A100-80GB and an Apple
MacBook M3 with MPS enabled.
\begin{table}[htbp]
    \centering
    \begin{tabular}{lccccc}
        \toprule
        \textbf{Device} & \textbf{Batch-Size} & \textbf{Model}  & \textbf{mean} & \textbf{min} & \textbf{max} \\
        \midrule
        CPU  & 32  & heron & 0.643 & 0.141 & 0.826 \\
        \midrule
        CUDA & 100 & heron & 0.030 & 0.026 & 0.099 \\
             & 200 & heron & 0.031 & 0.027 & 0.068 \\
             & 500 & heron & 0.026 & 0.026 & 0.027 \\
        \midrule
        MPS  & 50  & heron & 0.044 & 0.041 & 0.051 \\
             & 100 & heron & 0.060 & 0.041 & 0.091 \\
        \bottomrule
    \end{tabular}
    \caption{Inference runtime per image in seconds for the Heron model on various batch sizes. CPU=AMD EPYC 7763 (4 threads). GPU=A100 80GB. MPS=M3 Max, 40 cores, 64GB.}
    \label{tab:heron-runtime}
\end{table}

\label{app:appendix_c_runtime_overall}
Runtime performance of \textit{granite-docling-2stage-258M} compared to \textit{granite-docling-258M} benchmarked on \textit{DocLayNet} test set.

\begin{table}[htbp]
    \centering
    \caption{Comparative inference runtime (s/page) on \textit{DocLayNet} (NVIDIA A100, BS=1).}
    \label{tab:runtime_comparison_overall}
    \small 
    \begin{tabular}{lccc}
        \toprule
        \textbf{Model Configuration} & \textbf{Mean} & \textbf{Median} & \textbf{Std. Dev.} \\
        \midrule
        \textit{granite-docling-258M} & 2.26 & 1.96 & 1.52 \\
        \textbf{\textit{granite-docling-2stage-258M}} & \textbf{2.61} & \textbf{2.29} & 1.68 \\
        \midrule
        \textit{Abs. Overhead} & +0.35 & +0.33 & -- \\
        \textit{Rel. Increase} & 15.4\% & 16.8\% & -- \\
        \bottomrule
    \end{tabular}
\end{table}

\newpage
\subsection{Prompt Token Overhead}
\label{app:appendix_c_tokens}
Token-overhead statistics for layout-prior injection on the DocLayNet test split are summarized in Table~\ref{tab:heron-token-overhead}.

\begin{table}
    \centering
    \caption{Prompt token overhead from \textit{docling-layout-heron} on DocLayNet.}
    \begin{tabular}{l c}
        \toprule
        \textbf{Statistic} & \textbf{Tokens per page} \\
        \midrule
        Min & 0 \\
        Max & 590 \\
        Median & 74 \\
        Mean & 80.28 \\
        \bottomrule
    \end{tabular}
    \label{tab:heron-token-overhead}
\end{table}

\end{document}